\documentclass[sigconf]{acmart}

\usepackage[ruled,vlined,linesnumbered]{algorithm2e}  % added March 24
\usepackage{multicol} % added March 30
\usepackage{multirow} % added April 1

\AtBeginDocument{%
  \providecommand\BibTeX{{%
    \normalfont B\kern-0.5em{\scshape i\kern-0.25em b}\kern-0.8em\TeX}}}

\copyrightyear{2021}
\acmYear{2021}
\setcopyright{acmcopyright}\acmConference[UIST '21]{The 34th Annual ACM
Symposium on User Interface Software and Technology}{October 10--14,
2021}{Virtual Event, USA}
\acmBooktitle{The 34th Annual ACM Symposium on User Interface Software and
Technology (UIST '21), October 10--14, 2021, Virtual Event, USA}
\acmPrice{15.00}
\acmDOI{10.1145/3472749.3474771}
\acmISBN{978-1-4503-8635-7/21/10}

\begin{document}

\title{Hierarchical Summarization for Longform Spoken Dialog} 

\author{Daniel Li}
\authornote{Both authors contributed equally to this research.}
\email{daniel.li@columbia.edu}
\affiliation{
    \institution{Columbia University}
    \city{New York}
    \state{New York}
    \country{USA}
}

\author{Thomas Chen}
\authornotemark[1]
\email{chen.thomas@microsoft.com}
\affiliation{
  \institution{Microsoft}
    \city{Redmond}
    \state{Washington}
    \country{USA}
}

\author{Albert Tung}
\email{atung3@stanford.edu}
\affiliation{
  \institution{Stanford University}
    \city{Palo Alto}
    \state{California}
    \country{USA}
}

\author{Lydia B. Chilton}
\email{chilton@cs.columbia.edu}
\affiliation{
 \institution{Columbia University}
  \city{New York}
  \state{New York}
  \country{USA}
 }

\begin{abstract}
Every day we are surrounded by spoken dialog. This medium delivers rich diverse streams of information auditorily; however, systematically understanding dialog can often be non-trivial. Despite the pervasiveness of spoken dialog, automated speech understanding and quality information extraction remains markedly poor, especially when compared to written prose. Furthermore, compared to understanding text, auditory communication poses many additional challenges such as speaker disfluencies, informal prose styles, and lack of structure. These concerns all demonstrate the need for a distinctly speech tailored interactive system to help users understand and navigate the spoken language domain. While individual automatic speech recognition (ASR) and text summarization methods already exist, they are imperfect technologies; neither consider user purpose and intent nor address spoken language induced complications. Consequently, we design a two stage ASR and text summarization pipeline and propose a set of semantic segmentation and merging algorithms to resolve these speech modeling challenges. Our system enables users to easily browse and navigate content as well as recover from errors in these underlying technologies. Finally, we present an evaluation of the system which highlights user preference for hierarchical summarization as a tool to quickly skim audio and identify content of interest to the user.
\end{abstract}




\maketitle

\section{Introduction}  

Spoken dialog is a rich source of information; many media platforms frequently host discussions on important topics ranging from healthcare and diversity to economics and politics. Unfortunately compared to text, spoken dialog can be challenging to consume as it is slower than reading and difficult to skim or navigate. 
 Although people may be interested in a given topic, they may be unwilling to commit the required time necessary to consume long form auditory media given uncertainty as to whether such content will live up to their expectations. There exists a clear need to provide access to the information spoken dialog provides in a manner through which individuals can quickly and intuitively access areas of interest without investing large amounts of time.

An ideal solution would be to automatically summarize the content and distill it to its most interesting points, but this is problematic for three reasons. 
First, despite many advances in machine learning, Automatic Speech Recognition (ASR) and summarization are not yet mature enough to accomplish this. 
Second, there is a question as to whether the ASR transcripts and summaries can be trusted to be accurate, especially in the presence of informal language, minimal structure, and speech disfluencies. 
Third, what each user wants from a summary will differ based on their previous knowledge and expertise on the subject matter -- summaries are not one-size-fits all. This makes it difficult to provide training data for summaries that would be acceptable to a wide range of users, even if machine learning algorithms were perfectly accurate. 
\color{black}
We want to explore solutions that can leverage the strengths of machine learning, while overcoming many of its weaknesses.

\color{black}

We present a system that produces hierarchical summaries of spoken dialog that allow a user to browse and navigate the content to find things that are interesting to them. Hierarchical summarization allows users to first see a high level summary of the content and then drill into progressively longer and more detailed summaries - or listen to the raw audio itself. This approach addresses two key issues: 
\begin{enumerate}
    \item It allows users to be in control of what information they read at a high level and what information they consume in greater detail. 
    \item When machine learning (ML) models makes mistakes in ASR and summarization, users can quickly recover the ground truth.
\end{enumerate}

\color{black}
Although the typical approach to creating automated summarization systems requires training data that is difficult to obtain, our approach allows us to employ previously trained ML models recursively to generate shorter and shorter summaries. However, reusing models that were trained on different data requires careful model selection as well as novel algorithms to semantically segment the input text and thus output coherent summaries. 

This paper makes the following contributions:
\begin{enumerate}
    \item An end-to-end system that automatically generates hierarchical summaries of longform spoken dialog.
    \item A novel semantic segmentation algorithm that allows the reuse of existing machine summarization models rather than training a new one. 
    \item A user study demonstrating:
    \begin{enumerate}
        \item the system is 72\% accurate in producing condensed \textit{Short Summaries}.
        \item system hierarchical features enable users to recover their understanding of 98\% of summaries despite ASR and ML summarization model errors.
        \item the average time that users spent to reach an understanding of an audio recording was 27\% of the original audio length.
    \end{enumerate} 
    \item Qualitative findings about how people use \textit{Short Summaries} as navigational tools to help them "skim" audio and find the content most interesting to them. 
\end{enumerate}

\section{Related Work}

We discuss the four primary areas that our work builds on top of: (1) natural language processing (NLP) of audio and video to generate key points, (2) summarization of multi-party audio such as meetings and podcasts, (3) state of the art ASR and abstractive summarization, and (4) recursive summarization of complex content at varying levels of detail. We leverage several of the techniques used in both the user studies and the summarization works to create our system. 

\subsection{Using NLP to Generate Multimodal Interactions}

Researchers have developed models and systems to easily navigate through videos and movies by navigating to the video clip and allowing users to interpret content \cite{barnes2010video, goldman2006schematic, jackson2013panopticon}. However, these videos require users to search visual information in a video they may know little about and is inapplicable to pure audio files. To solve these issues, some researchers have employed summarizing key content in text as a means of helping users easily digest long-form content \cite{pavel2014video}, \cite{pavel2015sceneskim}. More recent work has adapted the use of hierarchical information to provide users with multiple levels of summarization and information \cite{truong2021automatic}. We build atop these systems targeting multi-party audio transcripts which pose novel challenges because these transcripts necessitate proper semantic segmentation to preserve meaning across speakers while simultaneously leveraging the usefulness of hierarchical information.

Still other work utilizes NLP to generate multimodal interactions such as images for video editing or even adding visuals to existing audio files \cite{xia2020crosspower, xia2020crosscast}. However, they rely on human-created transcripts, hurting the ability for the system to scale without automatic processes. Furthermore, visual representations only represent higher level abstract topics not the summarizations needed to represent the speaker.

\subsection{Summarization of Multi-Party Audio}

Creating meaningful summarizations from multi-party audio has been a difficult problem for researchers, often requiring hierarchical transformers and speaker segmentations to effectively retain information. Many of these papers, however, require full end-to-end training on transformers and even custom datasets \cite{zhu2020hierarchical, li2019keep, vartakavi2020podsumm, karlbomabstractive, zheng2020twophase}. Still others also employ graph-based summarization and coreferences to better summarize discourse \cite{xu2019discourse}. Meanwhile, current unsupervised abstractive summarizations do not utilize deep learning summarization modules and require the use of word graphs and ranking algorithms \cite{shang2018unsupervised}. These works focus on learning end-to-end summarization which is not practical across multiple domains. Instead, we focus on utilizing these summarization systems as part of a larger unsupervised abstractive system to generalize and reduce the overhead needed to deploy and scale such a solution. 

\subsection{Automatic Speech Recognition and Abstractive Summarization}

Automatic Speech Recognition systems (ASR) are used to transcribe audio (word recognition) into a source language transcript and have recently made relatively significant strides in terms of practical performance. Additionally, state of the art ASR \cite{google2021asr} is no longer constrained by vocabulary and remains relatively robust, encouragingly extending word recognition to topical domains and noisy audio. 

Text summarization techniques can be classified into two categories: abstractive and extractive. Abstractive summarization generates a new unique summary of text given a context whereas the extractive summarization “quotes” and concatenates relevant portions to compose into a summary. Because of spoken language noise effects in ASR transcripts, extracting transcript segments verbatim often leads to poor summaries. Therefore, we opt for the current state of the art abstractive summarization model, PEGASUS \cite{pegasus}, which is able to achieve much higher human-quality summaries. This is achieved by innovatively changing the pre-training process from standard word level masked language modeling, where models learn language conventions and syntax by predicting individually removed words within sentences, to sentence level masked language modeling, where entire sentences are removed and then recovered. This training process gives PEGASUS a high level of document understanding and helps to distill important information. Though promising, like most language models, it is important to note that PEGASUS is tailored towards specific benchmark datasets such as news or social media and that performance does not translate across different data domains, especially when applied to speech specific noise and disfluencies.

\subsection{Recursive and Hierarchical Summarization}
 Summarization of long complex material into recursively shorter and more tractable artifacts has been previously explored and found to provide an effective avenue for gaining useful comprehension of content \cite{zhang2017wikum}. Notably, this work showcased an interface displaying multiple summaries with varying levels of detail resulting in users having superior substantive recall and enabling non-linear exploration of the source material. However, this prior work employed crowd-sourced techniques to generate summaries and targeted solely threaded discussions typically found in forums. We build off these  findings by developing a novel system employing automatic summarization and speech recognition techniques to spoken dialogue in order to generate a similar hierarchical exploration of content without requiring human-in-the-loop summary generation. 
 
 The utility of hierarchical summarization has also been shown for multimodel instructional videos that use audio and video to demonstrate each instructional step \cite{truong2021automatic}. 
 By using computer vision, ASR, and domain-specific heuristics they  automatically group fine-level actions into coarse-level events (with summary text) that users can navigate at their own pace. 
 We build on these ideas by using machine summarization to provide multiple levels of summarization detail and allow users not only better navigation but also time savings in consuming media.

\section{Formative Study}
There has been much progress on machine learning models for natural language processing, including ASR and summarization. 
If possible, we want to use existing pretrained models as a component of our system to avoid the costly process of collecting longform summarized speech training data, as none exist or are readily available. This is particularly difficult for summarization because every user may want a slightly different summary.  
Moreover, there are two key problems: 
\begin{enumerate}
    \item ASR and summarization models are far from perfect and have inherent pre-existing challenges.
    \item Summarization models are almost always trained on text rather than speech data. If a text trained summarization model is deployed on speech data, there would be a data domain mismatch, leading to considerably degraded model performance. 
\end{enumerate} 

Compared to text, speech is far less structured - there are no topic sentences to rely on, speakers can stop mid sentence and backtrack their thought or never complete it, and coherency is challenging when multiple speakers are making different points simultaneously. Additionally, speech contains informal language and disfluencies such as hesitation and vocal fillers. These reasons heavily indicate that existing text-trained summarization models will perform very poorly on speech dialog.

To evaluate the practical performance of existing ASR and summarization models and determine which models to use as the basis of our system, we investigate the following criteria:
\begin{enumerate}
    \item \textit{Coherency}, are the final output summaries coherent? If this constraint is not met, the model is not usable. Aside from re-training and adapting a model towards speech data, we have no  tractable strategies for compelling model coherence.
    \item \textit{Information retention}, because output summaries are shorter and lossy, we check if they still retain salient information from the original passage. If a shortened summary does not contain useful or relevant information, it has no value.
\end{enumerate}

In the formative study, we identified three models that had various summarization properties and tested each model's reusability. Each model was applied to seven different recordings and an automatic evaluation score was computed to determine the quality of the summarization. To further substantiate each model's summarization, we check each model's performance with qualitative analysis.

\subsection{Evaluation Data}

We evaluate seven recordings of longform spoken dialog that span different topics, domains, and speech styles (Table \ref{tab:datasets}). The average length of the recordings is $32.5$ minutes and the average word count output from ASR is $5622$. Of these seven recordings, four are edited interviews from the NPR podcast "How I Built This", and 3 are unedited recordings from live events. Two are Bloomberg interviews regarding finance and one is a conversation about "How to foster true diversity and inclusion at work (and in your community)." These recordings were selected based on being content rich and of reasonable length. Information rich dialogue serves as a useful medium for this experiment by providing a sufficient density of information to showcase summarization. Additionally, choosing sources from the same producer reduces variance and provides a consistent structure for experiments. Finally, our experiment included both edited and unedited recordings of dialog to expose our system to both more coherent and structured conversations as well as free form dialogue.  
\begin{table*}
\caption{Dataset metadata used in formative study and final evaluation}
\label{tab:datasets}
\begin{tabular}{lllll}
\toprule
Transcript Name & Length & Word Count & Source & Edited? \\
\midrule
NPR: M. Night Shyamalan & 48 minutes  & 9184 words & How I Built This podcast & Yes\\
NPR: Chipotle & 48 minutes & 7847 words & How I Built This podcast & Yes \\
NPR: Health & 29 minutes & 5102 words & How I Built This podcast & Yes \\
NPR: Teach for America  & 22 minutes & 3909 words & How I Built This podcast & Yes\\ 
Diversity and Inclusion & 23 minutes & 4201 words & Recorded Ted Talk Interview & No\\
Bill Ackman on Economy  & 29 minutes & 5140 word & Recorded Bloomberg TV Interview & No\\
Ray Dalio on Economy   & 29 minutes & 3971 words & Recorded Bloomberg TV Interview & No\\
\bottomrule
\end{tabular}
\end{table*}

\subsection{Automatic Speech Recognition Model}
For word recognition, we use a state of the art ASR model, publicly available with the Google Speech-to-Text API. This system is already robust to a variety of domains and speech noise, while providing features such as diarization (speaker detection) and punctuation prediction. While we suspect the ASR component will not be a large contributing factor to poor summarization, we conduct a brief investigation on word recognition errors (word errors, i.e. homonyms such as \texttt{weather} compared to \texttt{whether}) as they could non-trivially impact downstream summarization performance. 

\subsection{Summarization Models}
\begingroup
\setlength{\tabcolsep}{4pt}
\begin{table}
  \caption{Model nomenclature where \texttt{\textbf{M}}$i$ indicates Model $i$, training data descriptions, and model maximum input and typical output sizes.}
  \label{tab:models_desc}
  \begin{tabular}{clll}
    \toprule
    Model & Domain / Fine-Tune Data & Max Words & Output Size\\
    \midrule
    \texttt{\textbf{M1}} & XSUM News / BBC News & 64 words & 1 sentence\\
    \texttt{\textbf{M2}} & News / CNN, DailyMail & 128 words & 3-5 sentences\\
    \texttt{\textbf{M3}} & Paraphrase / Quora, PAWS & 60 words & 1 sentence \\
  \bottomrule
\end{tabular}
\end{table}
\endgroup
For summarization, we investigate the current abstractive state of the art language model PEGASUS. While PEGASUS is noticeably improved over other summarization methods in terms of producing human level quality summaries, it requires fine-tuning onto domain specific summarization data. It is also important to note that a pre-trained only instance of PEGASUS is not normally used without modification; the pre-training procedure is different from summarizing and the authors focus solely on fine-tuned downstream summarization datasets. Appropriately, we select fine-tuned instances from \texttt{huggingface.co} \cite{hugg} that generate complete and grammatically correct passages (i.e. not a few keywords) and are still in considerably general domains (i.e. not a medical field model instance) to assess PEGASUS coherence and information retention. Model details are given in Table \ref{tab:models_desc}.

We begin by processing audio files to obtain raw ASR transcripts. However, because of the nature of longform dialog, the number of words per transcript greatly exceeds the maximum input length that \texttt{\textbf{M1, M2, M3}} can accept. Transcripts must be processed and split into manageable lengths. We naively segment the transcript in fixed $60$ word length segments set to \texttt{\textbf{M3}}'s maximum input length\footnote{We also experimented with increasing the input size to $128$ for \texttt{\textbf{M2}}, but still observed poor results (in fact, noise artifacts and incorrect model behaviors were more exaggerated than using $60$ word length segments)}. For example, if an input transcript segment had a total of $154$ words, it would be broken into a list of 3 individual segments, each containing $[60, 60, 34]$ words. To maintain evaluation consistency across all models, any evaluation involving naive fixed segmentation is set to $60$ words. These are then summarized by \texttt{\textbf{M1}}, \texttt{\textbf{M2}}, and \texttt{\textbf{M3}}, which are set to output summaries containing at most half of the original passage's words. 

\subsection{Heuristic Score}
We evaluate a summarization model's coherency and information retention using a heuristic score consisting of state of the art automated metrics in natural language processing. For coherence evaluation, we use a \texttt{BERTScore} \cite{zhang2020bertscore} between a reference ASR segment and a model generated summary (candidate input). This method correlates well with human evaluation and uses word level contextualized embeddings to capture dependencies and word ordering. For retained information, we use the cosine similarity between \texttt{Sentence Transformer} \cite{reimers2019sbert} embeddings of a reference ASR segment and a model generated output summary. A higher cosine similarity between the reference ASR segment and output summary suggests the summary captures the reference ASR segment's semantic content. The final heuristic is the simple average of the two and has a range of $[-1, 1]$. In practice, cosine distance based metrics used to determine similarities between word embeddings are positive, with a general range of $0-0.5$ for a weak correlation, $0.5-0.8$ for a moderate correlation, $0.8-1$ for a strong correlation, and $1$ for a perfect correlation \cite{jatnika2019word2vec}. As a sanity check, we observe a correlation of $1.0$ when we set the reference and candidate text inputs to be the same. Intuitively, as \texttt{\textbf{M1}}, \texttt{\textbf{M2}}, and \texttt{\textbf{M3}} outputs are still summaries, they will contain at least some semantic similarity to the reference ASR segment; therefore we expect to observe a somewhat moderate correlation ($0.5-0.6$) with our heuristic. After determining which model can be feasibly re-purposed, we use the heuristic score again to evaluate our method's impact towards improving summarization (Section \ref{sec:heuristic_2}). A more exhaustive discussion on heuristic score motivations, details, and limitations is given in the appendix, section \ref{sec:heuristic_score}. 

\subsection{Formative Study Findings}
We discuss Table \ref{tab:examples} throughout this section. It contains an example of the ASR transcript segment of one speaker in the "Diversity and Inclusion" recording and the corresponding summaries generated by the three models. Text is color coded to indicate shared regions between the ASR transcript and the summary.

\begin{table*}
\centering
\caption{Example summary generations on the "Diversity and Inclusion" dialog showing an instance of the ASR transcript for one speaker turn and errors frequently made by the three summarization models. Text is color coded to indicate shared regions between the input and output.}
\label{tab:examples}
\begin{tabular}{p{5.25cm}|p{6.1cm}|p{4.55cm}}
\toprule
ASR Transcript Segment & Summary & Description  \\\midrule
\multirow{3}{5.25cm}{\textcolor{blue}{opportunity is inclusion because you know, I have heard this stories so many times about there's you know}, \textcolor{magenta}{no Black Talent out there. No Latin} next talent for these particular roles. \textcolor{orange}{The talent is out there}. I will tell you that is underdeveloped because \textcolor{cyan}{I think we have spent more time trying to reach numbers}. Then \textcolor{olive}{we have changing our} \texttt{[$60$ words]}} 
& \texttt{\textbf{M1}}: \textcolor{red}{\textit{In our series of letters from African-American journalists, journalist and columnist Richard Lawson}} looks at the importance of \textcolor{blue}{inclusion} in the workplace.    
& \textbf{Hallucinations}, \texttt{\textbf{M1}} produces words and concepts (\textcolor{red}{\textit{italicized red}}) in the summary are not present in the given input. \\\cline{2-3}
& \texttt{\textbf{M2}}: \textcolor{blue}{opportunity is inclusion because you know i have heard the stories so many times} about theres \textcolor{magenta}{no black talent out there no latin next talent}. \textcolor{cyan}{i think we have spent more time trying to}
& \textbf{Speech noise sensitivity},  \texttt{\textbf{M2}} loses abstractive summarization ability and outputs unimportant sections. \\\cline{2-3}
& \texttt{\textbf{M3}}: \textcolor{cyan}{I think we have spent more time trying to reach numbers} than we should because there is \textcolor{magenta}{no black or latin} \textcolor{orange}{talent out there}.
& \textbf{Semantic Inaccuracy}, \texttt{\textbf{M3}} combines two different sentences into a semantically incorrect output.\\\bottomrule
\end{tabular}
\end{table*}

\subsubsection{Google Speech-to-Text Automatic Speech Recognition Quality}
We quickly and quantitatively evaluate the word error rate (WER) of the ASR system. Because Table \ref{tab:datasets} only consists of audio data and perfect transcripts (human transcribed) are non-existent, we benchmark ASR performance on a random subset of Ted Talks as they are somewhat similar in terms of speech and data structure to Table \ref{tab:datasets} and thus would likely be indicative of ASR performance. We find an average WER of $10\%$\footnote{This number should be treated as an upper bound as the human transcribed transcripts contain artifacts such as \texttt{"(Applause)"} or \texttt{"(Laughter)"}.}, slightly above the reported $6.7\%$ WER \cite{kim2019comparison}, and far below a usability constraint of $30\%$ \cite{gaur2016effects}.

As seen in the provided ASR Transcript example in Table \ref{tab:examples}, the ASR Speech-to-Text makes very few errors. However, rare words, unfamiliar phrases, or new words not yet encountered still degrade performance. For example, in the NPR: Chipotle dialog, \textit{"mise-en-place"} was mistakenly transcribed as \textit{"knees in place"}. In the "Diversity and Inclusion" dialog, \textit{"rectangle. Opening"} was mistakenly transcribed from \textit{"reckoning"}. 
Additionally, performance can fluctuate due to a variety of noising factors such as speech disfluencies, foreign accents, and audio recording quality. Although ASR makes few errors, they will propagate to downstream tasks and create challenges for generating a practical audio summarization system. 

\begin{table}[h]
  \caption{Automatic evaluation heuristic scores for various segmentation strategies. }
  \label{tab:model_eval}
  \begin{tabular}{clccc}
    \toprule
    Model Name & Segmentation Strategy & Heuristic Score \\%& \% Improvement \\
    \midrule
    \texttt{\textbf{M1}} & Naive Fixed Length & $0.61$ \\%& 36.7\% \\
    \texttt{\textbf{M2}} & Naive Fixed Length & $0.70$ \\%& 18.8\%\\
    \texttt{\textbf{M3}} & Naive Fixed Length & $0.68$ \\%& %21.4\%\\
  \bottomrule
\end{tabular}
\end{table}

\subsubsection{Summarization Model Quantitative Analysis}
Table \ref{tab:model_eval} gives the automatic evaluation heuristic scores for \texttt{\textbf{M1}}, \texttt{\textbf{M2}}, and \texttt{\textbf{M3}} ranging from $0.61-0.70$. Despite generating summaries on out of domain speech data, we can conclude that all the baseline language models can still reasonably function and retain a moderate amount of information with a summary containing at most half the words as the input ASR segment. Nonetheless, the the tight spread of the heuristic score range indicates a moderate correlation and merits further investigation into the re-usability of \texttt{\textbf{M1}}, \texttt{\textbf{M2}}, and \texttt{\textbf{M3}} to fully understand model behaviors.
While the heuristic score is telling, it is not a replacement for human level evaluation; it provides only a limited perspective into performance that is subject to intrinsic methodology constraints enumerated in Appendix \ref{sec:heuristic_score}.2. To get a sense of what types of errors the automatic summarization models are making and whether they could potentially be addressed, we studied various segments by hand. 

\subsubsection{Summarization Model Qualitative Analysis}
This style of evaluation was not formal; the errors were pronounced, ubiquitous, and immediately apparent. Such poor performance severely impeded practical usability and therefore did not necessitate a formal evaluation. Unfortunately, we observe that all three summarization models make frequent and substantial errors; however, \textbf{\texttt{M3}} stood out as containing problems that were addressable. 

\textbf{\texttt{M1}} produced summaries that contain frequent hallucinations \cite{maynez2020faithfulness} -- phrases or entities that appear to be semi-relevant but are not actually present in the underlying text. This can be attributed to its news based training data.  

For example, in Table~\ref{tab:examples} \texttt{\textbf{M1's}} summary contains the text ``African-American journalists'' and ``Richard Lawson.'' Neither of these entities are  mentioned in the input (or entire audio file). However, these entities are in \texttt{\textbf{M1's}} training data. This is a typical problem seen in language models when deployed on new data that is not encountered in training. Only recently, an attempt at fixing hallucinations has resulted in improved ROUGE precision and increased human preference \cite{zhao2020reducing}, but still requires additional dataset generation. These errors are in almost every summary produced by \textbf{\texttt{M1}}. Thus, fixing \textbf{\texttt{M1}}'s hallucinations would be nontrivial and require a new training dataset.

\textbf{\texttt{M2}} does not contain hallucinations but  
unfortunately it introduces many grammatical errors and performs especially poorly with regards to fluency: sentences trail off without finishing and summaries consist of concatenated phrases that may be individually sensible but holistically incomprehensible. Moreover, it fails to produce an abstractive summary and defaults to an extractive behavior; it mostly picked sections of the input rather than summarizing the entire input. This is likely because \textbf{\texttt{M2}} is trained to produce longer summaries than \textbf{\texttt{M1}}, and thus it is not forced to produce abstractive summaries. Reiterating Section 2.4, it is essential for a speech summarization model to be abstractive. These errors are frequently in summaries produced by \textbf{\texttt{M2}}. 

\textbf{\texttt{M3}} has more fluent text with no hallucinations. However, it makes an egregious error of misrepresenting the content. The transcript clearly states that `\textit{`The [Black and Latin] talent is out there,''} but the summary introduces a negation to say that the talent is not there. The root of this problem is that \textbf{\texttt{M3}} coerces two different segments into a semantically incorrect summary. These errors occur when multiple non-sequitur or different topics are provided as a single input. Because abstractive summarization generates words that are not necessarily present in the source input text, they require a high degree of content understanding of the underlying semantic information in the passage \cite{sss} to successfully generate a semantically faithful summary; a poorly segmented input containing multiple different concepts would be exceedingly detrimental towards a model's semantic comprehension.
Thus, \textbf{\texttt{M3}}'s resulting coherent and abstract summaries (albeit with contextual misrepresentation errors) signal that:
\begin{enumerate}
    \item A successful semantically accurate segmentation that can group similar topics and ideas together, while splitting dissimilar sentences into a separate chunk can improve a model's semantic comprehension, and transitively improve summary generation accuracy.
    \item The summary context's input accuracy issue is now reframed as a processing challenge that does not require changes to the model's architecture, re-training, or additional annotated training data.\color{black}
    \item \textbf{\texttt{M3}} is able to maintain its abstractive nature, which is essential to summarizing dialog due to speech disfluencies and other noise artifacts.
    \end{enumerate}
\subsubsection{Formative Study Key Takeaways}
Based on this exploration of the three models, we hypothesize that \textbf{\texttt{M3}} is the best one to build on top of and reduces the challenge of practical dialog summarization to a tractable problem. \textbf{\texttt{M1}} and \textbf{\texttt{M2}} errors are exceedingly difficult to correct  without significant amounts of specialized speech training data. \textbf{\texttt{M2}}'s marginally higher score over \textbf{\texttt{M3}} is immaterial given \textbf{\texttt{M2}}'s disfluency and incoherence\footnote{Refer to Appendix \ref{sec:heuristic_score}.2 for an explanation to why \textbf{\texttt{M3}} still achieves a comparable score to the other models.}.
Although incoherent topic grouping is rarely the case in written language where ideas are well-formed and presented in a manner that is optimized for ease of understanding, it is usually the norm in spoken language where topics shift over time as speakers react to the last thing that was said.
Concretely, if we can segment transcripts into semantically cohesive segments, creating easier inputs and facilitating improved summarization performance, \textbf{\texttt{M3}} may be an effective summarization tool. When errors do remain, there is the fallback of the user using the hierarchical browsing features to investigate surprising or suspicious claims to see if the summary is consistent with the text.

\section{System}

We present a system that produces hierarchical summaries of spoken dialog that allow the user to browse and navigate the content to find things that are interesting to them. Hierarchical summarizing allows users to first see a high level summary of the content and to then drill into progressively longer and more detailed summaries - or listen to the raw audio itself. 

 As shown by our formative study, pre-existing technology performance drastically suffers when applied to speech and is still considerably below the requirement for practical usage. Therefore, in addition to the borrowed pre-existing ASR system (Google Speech-to-Text API) and language summarization model (\textbf{\texttt{M3}}, paraphrasing adapted PEGASUS), we develop a method to identify semantically related segments of text that can be input into the summarization model, then merged back together to maximize coherent summaries. This process can be done recursively to get increasingly shorter and more abstractive summaries.
 
 The core technical novelty and contributions within our speech summarization framework are as follows:
\begin{enumerate}
    \item A novel segmentation algorithm that creates semantically similar input blocks from an input ASR segment in order to maintain conceptual cohesiveness
    \item A semantic hierarchical clustering algorithm that joins conceptually similar ideas for logical subsequent (recurrent) abstract summarization
\end{enumerate}

The inclusion of these procedures to the two-stage framework enables not only grammatical and semantic cohesiveness but also facilitates various levels of summarization detail:

\begin{enumerate}
    \item \textit{Long Summary}: Cleaned ASR Transcript. At this stage, a transcript’s disfluencies and noises are cleaned and presented according to conversational order or speaker turns.
    \item \textit{Medium Summary}: Moderately Detailed Summarization. Similar \textit{Long Summaries} are merged and further paraphrased, providing key concepts along with essential details.
    \item \textit{Short Summary}: High level Summarization.  Similar \textit{Medium Summaries} are further merged to provide the transcript’s salient ideas in more concise language.

\end{enumerate}

\subsection{Interface}

\begin{figure*}[h!]
\centering
\includegraphics[width=17cm]{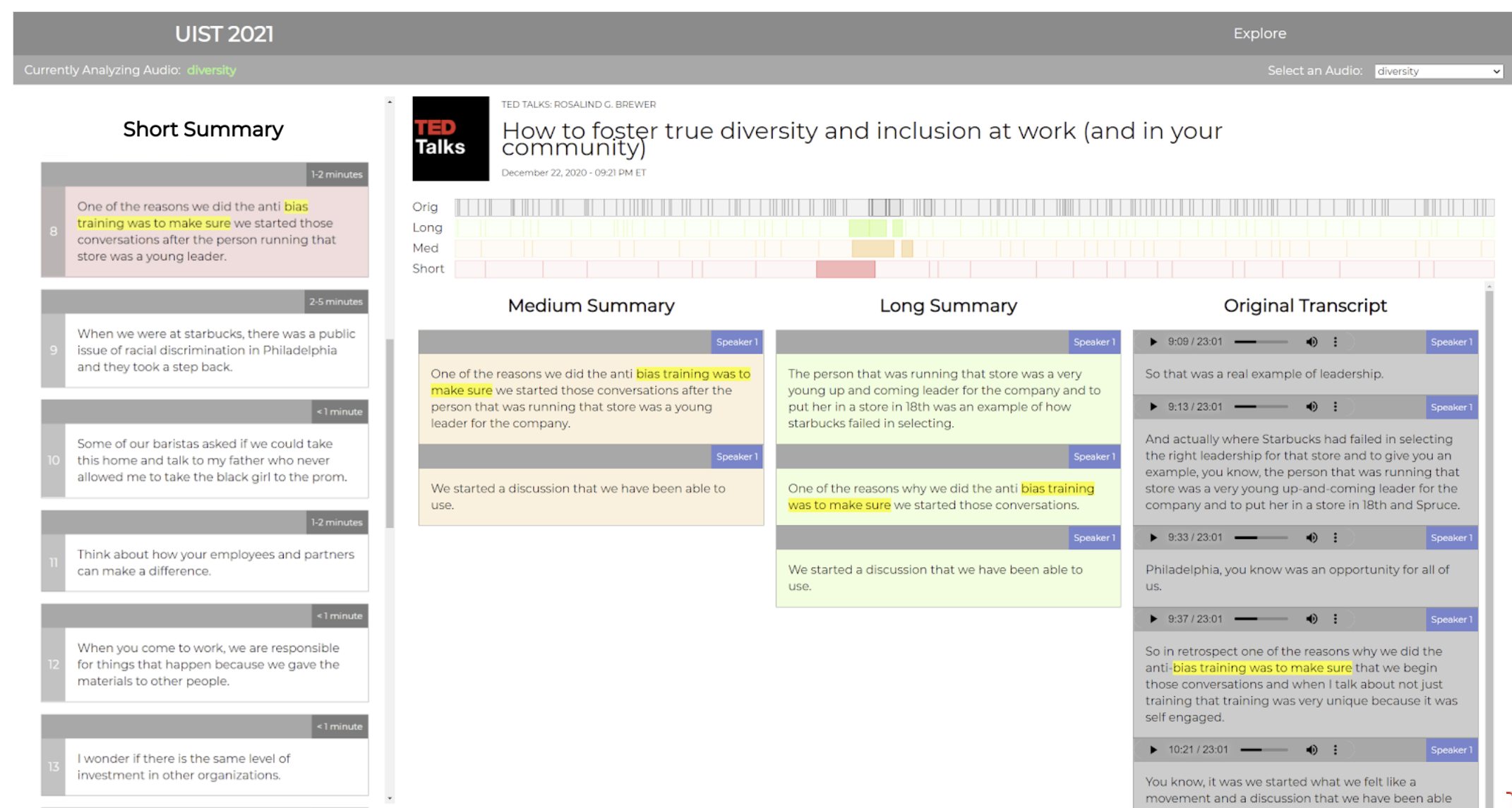}
\vspace{-0.2cm}
\caption{\textbf{System User Interface.} Example of the system's summarization display for each unique audio file. The left part of the interface contains several short summaries which, when clicked, displays the medium and long summaries along with the corresponding original transcripts and audio clips. The top part of the interface shows what part of the transcript each summary encapsulates information about. When the top part is clicked, users can navigate to any part of the summaries or transcripts.}
\label{fig:system_interface}
\end{figure*}

The interface (Figure \ref{fig:system_interface}) consists of three main sections: 
\emph{high level summary} column on the left, the  \emph{segment data view} in the middle and the \emph{timeline of segments} at the top of the interface.
Users explore the content by first browsing the \textit{Short Summary} column to get a high level overview of the content. 

Users may click a \textit{Short Summary} to see summaries of different length and additional levels of abstraction (\textit{Medium and Long Summaries}) as well as the ASR transcript, or elect to listen to the corresponding audio. 
Yellow highlighting shows a key phrase in the short summary and it's corresponding phrase in the other summaries and the ASR transcript to help orient readers as they move from reading the short summary to the other summaries of the same underlying text. 
The timeline of segments shows how all the summaries are aligned. The user can see some \textit{Short Summaries} that cover longer portions of the original transcript than others. Clicking on the timeline will take the user to the summaries of that section.

The interface was designed with two goals in mind: 

\subsubsection{Design Goal 1: Enable users to quickly identify useful information to them}
Presenting high level summaries to the reader allows them to quickly grasp a general idea of what is being said. However, simply reading \textit{Short Summaries} may not entirely satisfy the reader. By nature of being summaries, they may omit details that may be of interest. Additionally, the automatic summarization algorithms are imperfect and sometimes present summaries that are more vague than a user would prefer.
However, the purpose of the \textit{Short Summaries} are not necessarily to fully summarize, but to allow the user enough information scent \cite{info_foraging} to decide if they want more detail. If they want more detail, they can use the hierarchy of summaries to read \textit{Medium} or \textit{Long Summaries}, read the ASR transcript, or listen to the underlying audio. 

\subsubsection{Design Goal 2: Support error recovery}
As both automatic speech recognition and summarization may produce errors at various stages of the system, the interface provides multiple tiered layers of information for users to fall back on in order to recover comprehension of any given set of summarization data in the event that either a portion of the transcribed audio or summaries has erroneous text.

Listening to the raw audio will provide the full information a user should need to recover from confusion or loss of comprehension due to an ASR or summarization error. However, listening to audio takes longer for most people than reading text. If users wants near-full fidelity information in a form they can read (or scan), they can refer to the ASR transcript. Many find transcripts of dialog difficult to read because of the informal language and speech disfluencies. Thus, users may find the \textit{Long Summaries} easier to read - they retain almost all the information of the ASR transcript, but the text is cleaned up to remove these artifacts of speech. Users who want more actual summarization can refer to the \textit{Medium Summary}. 

By presenting users with these options for recovering from ASR and model errors, users can decide how much time and effort they want to put into recovering from the error. However, there is a possibility that offering users multiple options could provide a negative experience by overwhelming them with choices. Over time, we expect users will become familiar with the nature of each level of detail and get a sense of which option to select. This is an issue we address in the evaluation section.

\subsection{Summarization Algorithm and Implementation}

\begin{figure*}
\centering
\includegraphics[width=15cm]{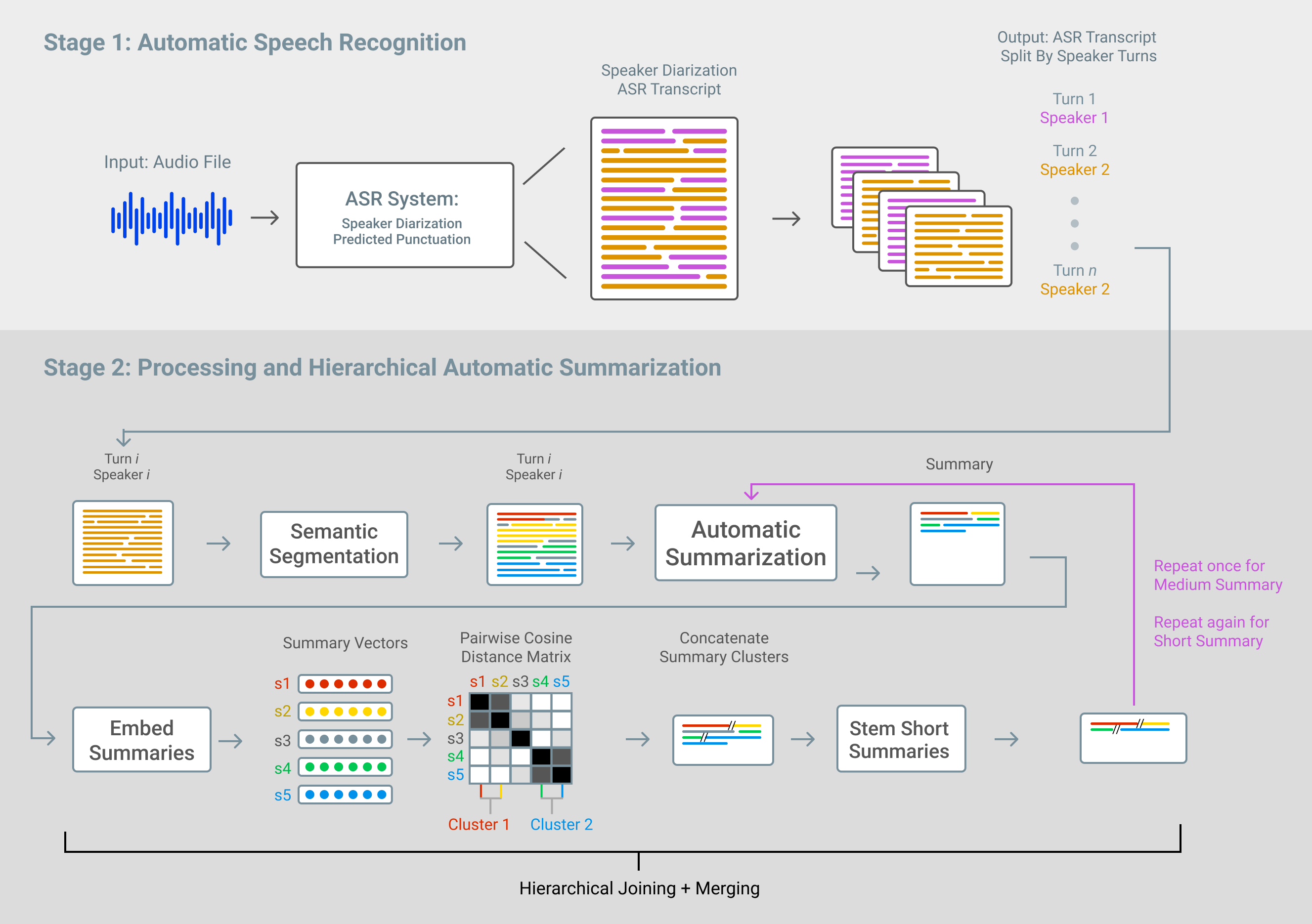}
\vspace{-0.2cm}
\caption{\textbf{Summarization Generation Pipeline.} Our system enables the conversion of audio files to multiple tiers of summarization. In the first stage, we convert the audio file into a speaker-segmented and punctuated transcript and process the transcript, split by speaker turns. In the second stage, we take each speaker turn and cluster conceptually similar summaries via semantic segmentation. Each cluster's summaries are joined (concatenated) based off of semantic similarity. We remove small summarizations and then repeat the summarization and merging process to obtain the \textit{Medium} and \textit{Short} summaries.}
\label{fig:pipeline}
\end{figure*}

Figure \ref{fig:pipeline} shows the steps through which an audio file is recurrently processed to obtain different levels of summarization (\textit{Short, Medium, Long}). 
At a high level, the system segments an ASR transcript and iteratively summarizes previously combined conceptually similar segments to obtain increasingly abstract summaries while preserving semantic meaning. 

Stage 1 (Fig. \ref{fig:pipeline}) of the system employs ASR to create a speaker diarized transcript of the input audio file. These speaker turns are further processed by a semantic segmentation algorithm which divides a given speaker turn into chunks of semantically related sentences. The now refined speaker turns are iteratively given as inputs into stage 2 (Fig. \ref{fig:pipeline}) of the system where processing and hierarchical automatic summarization occurs. After each speaker turn is individually summarized, its summaries are embedded which are then used to cluster sentences of the summary. Clusters are concatenated (merged) and shorter summaries, which generally contain little salient information, are stemmed. The first resulting summary of this iteration through the pipeline represents a \textit{Long Summary}. This \textit{Long Summary} is fed back into the the automatic summarization model and follows the same embedding, hierarchical clustering, and stemming steps once more to generate a \textit{Medium Summary}. One further cycle using the \textit{Medium Summary} yields a \textit{Short Summary}.

Details of each system component are as follows:

\textbf{Automatic Speech Recognition.} We begin by using the Speech-To-Text Google API to obtain transcripts with speaker diarization and predicted punctuation for initial sentence boundaries. Speaker turns are alternating blocks of text separated by changes in speaker; they provide a very coarse starting point for transcript segmentation. Speaker turns frequently discuss multiple different ideas and may result in a long monologue before another speaker interjects.

\begin{figure*}[h!]
\centering
\includegraphics[width=17cm]{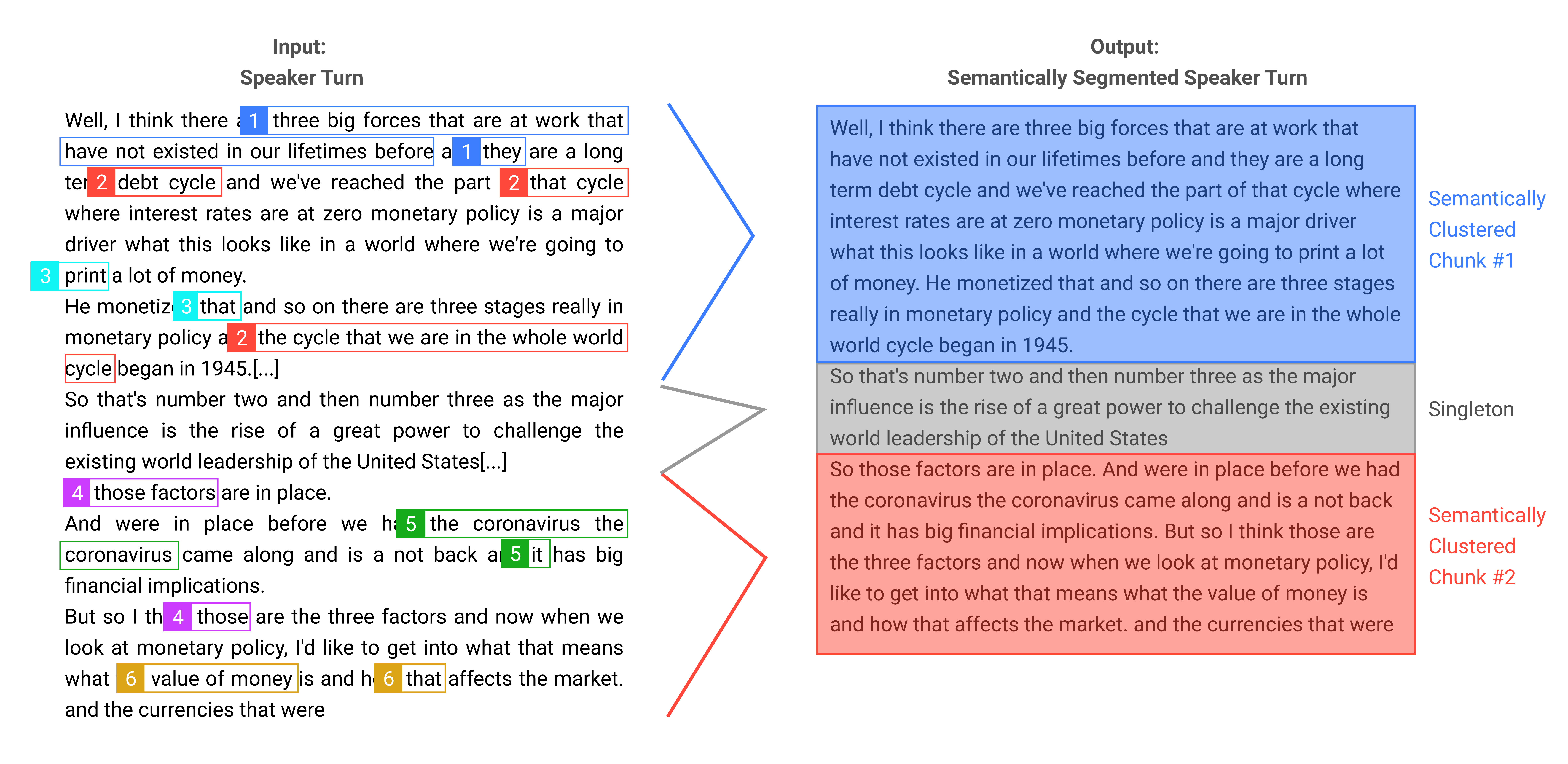}
\caption{\textbf{Alg 1. Semantic Segmentation} Example of one speaker turn input into our coreference resolution algorithm. On the left, the coreference tags generated from the AllenNLP coreference resolution module are shown with six different references highlighted. Our algorithm groups sentences with references into semantic chunks with a minimum limit of references and words so that the semantic chunks are still meaningful. In Semantically Clustered Chunk \#1 (blue), the first sentence includes three different references (1,2,3), of which two (2,3) are still used in the second sentence, hence why it is included. The singleton contains no references and is segmented out. In Semantically Clustered Chunk \#2 (red), the first sentence contains two references (4,5) of which one (4) is in the second sentence. Although there is a new reference (6), the reference terminates within the second sentence so no further sentences are added.}
\label{fig:coref}
\end{figure*}

\textbf{Coreferenced Semantic Segmentation}. To understand why we employ coreference resolution \cite{soon2001machine} and speaker shifts to semantically link sentences together and correct poor segmentation, we must recognize two linguistic tendencies: 

First, unlike written prose, conversation can be far more vague; nouns and objects, herein referred to as entities, are only initially mentioned and then sporadically referenced, while all other mentions are pronouns (it, s/he, they, etc.). Generic topic modeling of dialogue performs poorly due to the nature of conversations, since conversations have both local and global topic structures that have weak signals in conversation \cite{takanobu2018weakly}. Individuals can talk about a topic using specific references, but a model not trained to recognize these topics could fail to recognize the boundaries of the topic effectively. To eliminate a dependency on custom training data, we instead choose to identify expressions that refer to the same entity using coreference resolution. To employ this technique, our algorithm seeks to group sentences spanned by an entity, which is defined as the sentences contained by the start and end of the entity. This method of using coreferenced entities to model text has historically been shown to be successful \cite{stoyanov-cardie-2006-partially} \cite{witte2003fuzzy} and is still used in current state of the art models \cite{xu2019discourse}.
We use the publicly available Allen NLP API \cite{gardner2018allennlp} for state of the art coreference resolution. We directly refer to variables in the pseudocode for Algorithm \ref{sec:alg1}, given in appendix section A.\ref{sec:alg1}, in the following walk through of the process. 

Second, speaker shifts may begin relatively different concepts \cite{MAYNARD+1980+263+290} and repeated references to the same entity indicate the same concept is still being discussed. Sudden changes in speakers can be correlated with topic boundaries \cite{galley2003discourse} and this concept serves as a common segmentation approach in NLP \cite{zhu2020hierarchical}. All iterative instances of the algorithm on each speaker turn segment $S_i$ are therefore independent of each other.

Here we walk through a single speaker turn pass of Algorithm \ref{sec:alg1}. Speaker turn $S_i$ contains sentences $s$ that were previously divided by ASR predicted punctuation. For each sentence $s$, we first obtain the coreferenced entity $c_{min}$ (\texttt{line 13}) that maximally spans the current sentence $s$ and future consecutive sentences $s \in S_i$ (\texttt{lines 4:10}), subject to constraints. This denotes the start and end pointers ($e_0, e_f$) of the current semantically similar chunk, $\mathbf{P}$ (herein referred to as cluster, \texttt{line 13}).  Sentences contained by $\mathbf{P}$’s span $e_0, e_f$, are assigned to $\mathbf{P}$ (\texttt{lines 19:20}). If sentence $s$ contains another entity that spans further than $\mathbf{P}$’s current end pointer, $e_f$ is updated to the sequentially higher indexed sentence (\texttt{lines 21:23}). When $\mathbf{P}$’s span is exhausted and cannot be further extended, the algorithm begins a new cluster $\mathbf{P}$. Sentences that contain no entities are singleton clusters.

We restrict the entity span to at most $m=100$ words and require each valid span to contain at least $p=3$ mentions (coreferences).\footnote{We empirically observed that entities below these requirements had low relevance to the underlying concept.} We also do not consider \texttt{“I”} and \texttt{“me”} entity references since these references do not indicate a semantic change. Figure \ref{fig:coref} demonstrates an instance of the sentence entity spans procedure.

\textbf{Summarization Model.} We opt to reuse the paraphrasing \textbf{\texttt{M3}} instance of PEGASUS. The implementation of \textbf{\texttt{M3}} was taken from \texttt{huggingface.co}, using checkpoint \texttt{Tuner/007}. We also make the key observation that multiple recurrent forward passes of \textbf{\texttt{M3}} (independent of \texttt{Short, Medium, Long} heirarchical summarizations) removes speaker disfluencies and other speech artifacts of low importance. 

The tradeoff for increased robustness towards speech noise and artifacts is also inherently found in \textbf{\texttt{M3}}'s paraphrasing nature; \textbf{\texttt{M3}} struggles to reason out semantically different ideas and suffers substantially from contextual errors (Table \ref{tab:examples}, \textbf{\texttt{M3}}). However, when ASR transcripts are preprocessed with Algorithm \ref{sec:alg1}, our framework is able to generate not only cohesive and semantically logical summaries, but also achieve practical accuracy.

\textbf{Hierarchical Concept Clustering and Merging.} The next challenge is to determine which of the previous level's summaries to concatenate for further abstract summarization (appendix section A.\ref{sec:alg2}). Recall that semantically similar summaries must be joined or the model output can be factually incorrect (Table \ref{tab:examples}, \textbf{\texttt{M3}}) as abstractive summarization requires a high degree of semantic understanding of the underlying input passage. As a means to compare summary content similarity, we first use \texttt{Sentence Transformer} \citep{reimers2019sbert} to individually embed summaries (still contained in their own speaker turns $\mathbf{S}$) $s \in {S}_{i=0}^{n}$, into vectors in a semantic space. By transforming the text segments into vector representations, we can now quantitatively compare their similarities via cosine distance. Summaries within each speaker turn ${S}_i$ are then merged through usage of a pairwise cosine distance matrix for hierarchical (agglomerative) clustering. Merges are done within speaker turns to enforce a proximity constraint of only merging local summaries due to the long lengths of ASR transcripts. 

Next, identified summary clusters are sequentially concatenated and concatenated summaries containing $5$ or fewer words are stemmed. We observed that summaries which are not merged and contain few words are very frequently speech artifacts that contribute no value. Pseudocode is given in Algorithm \ref{sec:alg2}.

\color{black}
\subsubsection{Coreferenced Semantic Segmentation Effectiveness on Long Summaries}
\label{sec:heuristic_2}
\begin{table}[h]
  \caption{Automatic evaluation heuristic scores for various segmentation strategies on \textit{Long Summaries} compared to \textbf{\texttt{M3}} using Coreferenced Semantic Segmentation.}
  \label{tab:full_model_eval}
  \begin{tabular}{clcc}
    \toprule
    Model  & Segmentation Strategy & Heuristic Score \\
    \midrule
 \texttt{\textbf{M3}} & \textbf{Coreferenced Semantic} &  \textbf{0.83}\\\hline
    \texttt{\textbf{M1}} & Naive Fixed Length & 0.61  \\
    \texttt{\textbf{M2}} & Naive Fixed Length & 0.70 \\
    \texttt{\textbf{M3}} & Naive Fixed Length & 0.68 \\
  \bottomrule
\end{tabular}
\end{table}

To determine if semantic segmentation is effectively grouping information for \textbf{\texttt{M3}}, we evaluate the core coreferenced semantic segmentation algorithm, Alg. \ref{sec:alg1} on \textit{Long Summaries}.
Specifically, we use our heuristic score to \textbf{only} evaluate \textit{Long Summaries}, as subsequent (\textit{Short, Medium Summaries}) evaluation using a previous longer summary as input would induce a circular dependency due to \texttt{Sentence Transformer}'s usage in Algorithm \ref{sec:alg2} and the heuristic itself. 
Because \textit{Short, Medium Summaries} use \texttt{Sentence Transformer} to determine which input passage segments (that would ultimately become a circular reference transcript) to semantically include for summarization, their outputs would likely score artificially high due to the heuristic's intrinsic incorporation of \texttt{Sentence Transformer}. 

We find that \textbf{\texttt{M3}} with semantic segmentation obtains a heuristic score of $0.83$, a $0.13$ improvement over the best naively segmented model, \textbf{\texttt{M2}}, suggesting Alg. \ref{sec:alg1} is effective and facilitates increased summarization performance. The marked improvement is important as mediocre initial summarization (\textit{Long Summaries}) would lead to poor downstream hierarchical summaries (\textit{Short, Medium Summaries}).

\section{Evaluation}

We performed user studies to evaluate the following:
\begin{enumerate}
    \item Human assessment of the quality of \textit{Short Summaries}.
    \item The system's ability to help users recover from errors in summaries.   
    \item The amount of time users saved by using our system when used in an unconstrained setting with their own browsing styles and comprehension goals.
\end{enumerate}

Additionally, we present qualitative findings on how and when people would use the system to find interesting information in spoken dialog.

\subsection{Methodology}We recruited $10$ recent university graduates 

from diverse professions ($5$ women, average age $= 26$ ) for our study. Each study lasted $1.2$ to $2$ hours and averaged $1.5$ hours; subjects were paid $\$20$ per hour for their time. 
To begin, participants were provided with a scenario where a dialog summarization tool would be potentially useful: \textit{imagine you get an email from a friend or colleague about an exciting interview on YouTube about “Diversity and Inclusion in the Workplace.” It’s 23 minutes long, you’re not sure if you want to commit to watching the whole thing, but you want to know if there’s anything new or interesting in it. We’re trying to help people explore audio clips to find key takeaways.} They were also informed that the summaries were generated by an AI and might be imperfect.

Participants were then given a link to the interface with the audio for "Diversity and Inclusion" loaded in. During the warm-up, we explained the different UI components, \textit{Short, Medium,} and \textit{Long} levels of summarization, the original transcript, and the media button to play and scan the corresponding original audio section. Figure \ref{fig:system_interface} is an example of what the user would see. To familiarize users with the system, we instructed them to read the first \textit{Short Summary}, its corresponding \textit{Medium Summary}, \textit{Long Summary}, and ASR transcript segment, as well as to play the audio segment.

After the warm-up, participants were asked to perform three tasks: 
\begin{enumerate}
    \item \texttt{Short Summary Quality  Assessment}: assess \textit{Short Summary} quality for two audio files ("Diversity and Inclusion" and one of their choice). For \texttt{Short Summary Assessment}, participants were asked to think aloud so we could understand how participants built an intuition and what their interpretation of the system was like. We asked participants to  rate each \textit{Short Summary} for two things: 1) grammatical correctness and 2) semantic comprehensibility. Users answered "yes" or "no" to each question.
    For semantic comprehensibility, we ask them whether they were able to understand the \textit{Short Summary} and if it matched the corresponding audio segment's content.  \color{black} Users were able to check for semantic meaning by comparing against \textit{Medium, Long Summaries}, original ASR transcripts, and audio. 
     \item \texttt{Short Summary Error Recovery}:
    if participants were confused on a \textit{Short Summary}'s meaning, they were asked if 1) they could regain comprehension of the \textit{Short Summary}'s meaning and 2) what system features they used to recover the meaning. Participants were allowed to spend as much time as they needed to rate all the \textit{Short Summaries} for the two audio files and were encouraged to think aloud as much as possible.
    \item \texttt{Practical Usage Assessment}: use the system as they would in an every day situation. Participants were asked to choose the audio file that interested them most from the 5 remaining audio files in Table \ref{tab:datasets} and to use the system to find interesting information in that dialog quickly. \color{black} They used the system without any restrictions and without thinking aloud. We timed participants' usage and observed their browsing strategies.
\end{enumerate}

We concluded the study with a semi-structured interview about their experience using the system. 

\subsection{Results}
\label{sec:results}

\subsubsection{\textit{Short Summary} Accuracy} During the study participants rated a total of 556 \textit{Short Summaries}. 
The overall average accuracy across all users and recordings was $71.4\%$ (see Table \ref{tab:main_accuracy}). The overall accuracy includes both grammatical correctness ($84.9\%$) and semantic comprehensibility ($75.9\%$). Overall accuracy is a measure of how many \textit{Short Summaries} had any kind of error (either grammatical or semantic).

\begin{table}[h]
  \caption{Average \textit{Short Summary} accuracy and standard deviations.}
  \label{tab:main_accuracy}
  \begin{tabular}{l|l}
    \toprule
    Criteria & Accuracy \\
    \midrule
    Grammatical Correctness & $84.9\pm 5.1$\% \\
    Semantic Comprehensibility & $75.9\pm 4.8$\%\\
    \textbf{Overall Accuracy} & $\mathbf{71.4\pm4.9}$\%\\
  \bottomrule
\end{tabular}
\end{table}

An overall accuracy of $71.4\%$ means that many \textit{Short Summaries} can be read and understood without any issues. 
The system's grammatical correctness is reasonably high ($84.9\%$), but the system's semantic comprehensibility is lower ($75.9\%$).
Generally, grammatical errors are not detrimental to user experience because most grammatical errors do not distort the meaning of the sentence. However, poor semantics often requires users to  investigate further to comprehend the meaning \cite{kaschak2000constructing}.
In this study, to fully comprehend every \textit{Short Summary}, users would need to investigate semantic errors for 1 in 4 \textit{Short Summaries}.

Because of the current state of machine summarization, we were not expecting the \textit{Short Summaries} to be perfect. However, with $71\%$ accuracy, we are encouraged that a usable and valuable system can be built in presence of errors. The subsequent evaluations are focused on whether the the system can help users achieve their goals despite these errors.

\subsubsection{Short Summary Error Recovery} When a participant encounters a confusing \textit{Short Summary} during the \texttt{Short Summary Quality Assessment} task, we evaluate whether the user can recover regain holistic text comprehension by using the interface. In total, there were 140 \textit{Short Summaries} with unclear meaning, and users recovered from  
$\mathbf{92.9\%}$ of them.
This recovery rate is high -- the interface allows them to recover from all but $\mathbf{7.1\%}$ of them.
This indicates that although  \textit{Short Summaries} contain errors, the system can still allow users to have full comprehensions with some extra effort --  the post-recovery success rate is nearly perfect: 98.2\%.

\begin{table}[h]
  \caption{Distribution of the hierarchical levels users explored in order to recover from an inaccurate \textit{Short Summary}.}
  \label{tab:recoveries}
  \begin{tabular}{l|l}
    \toprule
    Required Hierarchical Traversal Level & Fraction (\%) Used \\
    \midrule
    \textit{Medium Summary} & $11.5\%$ \\
    \textit{Long Summary} & $19.2\%$ \\
    ASR Transcript Segment & $40.8\%$ \\
    Audio Segment & $20.8\%$ \\
    Querying Neighboring Segments & $6.9\%$\\
  \bottomrule
\end{tabular}
\end{table}

The hierarchical summarization features of the interface were designed to help users recover from errors quickly and easily.  We wanted to know to what degree participants used these features during recovery. We found that participants used all the hierarchical summarization features to some degree. Participants had two main styles of using the summarization: either they traversed down the hierarchy in order (from most summarized to least summarized forms of the semantic chunk or skipped to their preferred source of information.  Table  \ref{tab:recoveries} shows the breakdown of how often each feature was used. Participants used \emph{Medium Summaries} a small amount (in $11.5\%$ of recoveries) and used \emph{Long Summaries} more (in nearly $20\%$ of recoveries). However, participants used the 
ASR Transcript Segment the most (for $40.8\%$ of recoveries.)
This behavior is explained by users noting how \textit{Medium, Long Summaries} were either too similar or contained the same semantic errors as \textit{Short Summaries}. As a result, users defaulted to reading the ASR Transcript Segment more often.

There are some instances where the transcript and summaries are insufficient for error recovery. In $20.8\%$ of recoveries, users chose to go back to the audio segment. Although audio takes more time to listen to, the audio contains information that the transcript does not - it contains emphasis and tone of voice (as well as avoiding any errors in the transcript). In a small number of cases ($6.9\%$) users chose to read neighboring segments to recover from an error. This is almost always because users needed more context that lay outside of the current semantic chunk to recover full comprehension. 

Lastly, some participants noted that some of these errors may not be possible to recover from  because the underlying audio content was difficult to understand or incoherent.

\color{black}

\subsubsection{Time Savings} 
In the \texttt{Practical Usage Assessment}, when participants used the system at their own pace (without thinking aloud or rating tasks), we found that on average, the 
time participants spent was $\mathbf{27.1\%}$ of the  original audio time to reach a level of understanding that they were satisfied with (Table \ref{tab:times}). 
The fastest user spent only $6.9\%$ of the original audio time, and the slowest spent $75.9\%$ of the original audio time, but most spent between $20\%$ and $30\%$ of the original audio time. We believe the system presents a sizable time savings. Although users do have strategies for processing audio faster (such as listening at 1.25x speed or skipping the first minute of the audio), these strategies are unlikely to provide dramatic speed ups and often lack the control and freedom that our system provides.

How much time users spent varied with what level of detail they wanted to understand. The two users with the highest time percentage ($75.9\%$ and $45.5\%$) were looking for details.  The other users took much less time and wanted a cursory understanding of the material (Table \ref{tab:times}) such as getting a broad gist of the conversation, wanting a few key takeways, or wanting only a specific piece of information. See Section \ref{sec:qe} for more details. 

\begin{table}[h]
  \caption{Individual participant times using the system as percentage of original audio length alongside intended browsing style. }
  \label{tab:times}
  \begin{tabular}{ll|c}
    \toprule
    Participant & Self-Declared Style Browsing & \% of Audio Time \\
    \midrule
    $P_1$ & Detailed & $75.9\%$ \\
    $P_2$ & Cursory & $12.1\%$ \\
    $P_3$ & Cursory & $14.9\%$ \\
    $P_4$ & Cursory & $20.7\%$ \\
    $P_5$ & Cursory & $20.8\%$ \\
    $P_6$ & Cursory & $6.9\%$ \\
    $P_7$ & Cursory & $25.9\%$ \\
    $P_8$ & Moderately Detailed & $24.7\%$ \\
    $P_9$ & Cursory & $24.3\%$ \\
    $P_{10}$ & Moderately Detailed & $45.5\%$ \\\midrule
    $P_{average}$ & - & $27.2\%$ \\
  \bottomrule
\end{tabular}
\end{table}

\subsubsection{Qualitative Evaluation}
\label{sec:qe}
During the semi-structured exit interview, participants were asked about their experience using the system, use cases where they would or would not consider using it in their life, and potential improvements to the system.

While browsing for interesting nuggets of information, users sometimes found the \textit{Short Summary} sufficient, but often leveraged hierarchical summarization features to dig further. In the podcast discussing Teach For America during Covid, $P_7$ found this \textit{Short Summary} to be interesting on its own: “15 to 16 million children don't have access to broadband internet.” Likewise, in the podcast on Health, $P_8$ found this \textit{Short Summary} interesting without reading further: “You can prevent systemic bias by hiring nurses who speak Spanish and are bilingual.” However, $P_2$ selected Ray Dalio’s interview and found the first \textit{Short Summary} intriguing (“There are three big forces at work have have not existed before”) but had to read the transcript to discover what the three forces were. 
Similarly, $P_9$ selected Bill Ackman’s interview and was intrigued by the \textit{Short Summary}: “The uncertainty of the future can affect the model that analysts use to value securities.” $P_9$ said: “I found this as a thesis statement and read more into it.”
Although \textit{Short Summaries} may be sufficient, that is not always the case and as a result it is critical that \textit{Short Summaries} provide good information scent to indicate to users when to use hierarchical features to investigate further. 

Users reported they would consider using a tool like this for media they considered “condensible.” They mentioned news, YouTube videos (particularly reviews), interviews, and podcasts as media that could be condensed. Users also stated they would prefer to use the tool for topics they were curious about, but “didn’t want to spend too much time on" ($P_7$).
One such use case was if a friend suggested they listen to a long audio file and they “don’t wanna be rude” ($P_7$). 
Another use case presented was for situations when a given topic is familiar, but the presentation could give background that could be condensed ($P_2$). Finally, a third case users noted is when they wanted specific information from an audio recording, such as "learning what a company does in interviews with CEOs" ($P_{10}$). 6 out of 10 participants said they would not use the tool for detailed or technical topics, particularly if they were responsible for learning the material at work or school. Additionally, they would not use it for personal things they were deeply interested in because “I’d want to read those in detail” ($P_2$) or for fictional narratives where there is pleasure in enjoying the flow of the story rather than extracting information.
Clearly, this is not a tool for all use cases. Similar to how people wish to skim text using a tool, our system can allow users to “skim” audio. 

Throughout the experiment users frequently relied on their own knowledge to guide them towards exploring content in more detail. $P_5$ works in the medical field and was intrigued by a \textit{Short Summary} in the Health dialog: “There was a spike in demand for specific types of nurses.” $P_5$ wanted to know what those specific types were. The \textit{Medium Summary} did not contain that information but the \textit{Long Summary} did - ICU nurses and ED (emergency department) nurses were the two specialties named. 
$P_{10}$ was familiar with “How I Built This” podcast and was specifically interesting in understanding the “pain points with building the company.” He spent most of his time in the middle of the interview because he knew from the structure of the podcast that the information would probably be there.
$P_4$ was already familiar with finance and with Bill Ackman’s philosophies, but the tool was useful to him as he skimmed the \textit{Short Summary} to see if there would be anything new and interesting, given his background. 
For users with background knowledge, tools which provides user control and freedom enable more efficient navigation in order to locate valuable information.

The hierarchical features were more useful for some dialog than for others. 8 of 10 users of the Diversity dialog did not mention reading \textit{Medium, Long Summaries}, and always went to the transcript. However, 5 of 5 readers of Ray Dalio dialog used the \textit{Medium  Summaries}. This is likely due to the nature of the underlying audio and the quality of the summaries. Ray Dalio tends to speak in structured and organized paragraphs creating longer but structured ASR transcript segments. This created enough of a distinction between \textit{Short} and \textit{Medium Summaries} where \textit{Medium Summaries} contained a good balance of interesting information while retaining an attractive length. Meanwhile, the \textit{Medium, Long Summaries} of the Diversity dialog did not add enough information causing users to read ASR transcript segments to obtain desired additional details. As these factors are difficult to control for and user dependent, the solution of presenting summaries at multiple levels of granularity was successful. 

\textit{Short Summaries} were imperfect, but users found strategies to recover understanding of the underlying material. A common complaint about the tool was use of ambiguous pronouns in the \textit{Short Summaries}. For example,
in the Chipotle interview the \textit{Short Summary} says. “It's hard to say what he is like because he was an amazing visionary.” Here “he” refers to the CEO’s mentor - a head chef at a famous restaurant, but users had to read the transcript to find this information. A related complaint is that “the short summaries were disconnected from each other” ($P_7$). Summarization often removes segues and other transitional elements in order to surface meaning. However, this provides a disjointed experience for users and requires them to “rewind a little bit” ($P_2$) to recover context or flow. When using the hierarchical summarization features to recover users found “Word per word highlighting indicates where it was to quickly see the segment to read to resolve” ($P_{10}$). This type of consistency across the interface makes information easier to scan.
In future work, we could explore ways to make the \textit{Short Summaries} flow better, such as resolving pronouns and linking related information across the \textit{Short Summaries}. 

\section{Limitations and Future Work}
Users generally found the system useful, though there are several ways the underlying technology could be improved to provide better summaries and to generalize to more types of dialog.
\subsection{ASR Limitations}

Although ASR works well for the two-person, studio-recorded interviews in this system, it still has many limitations. ASR may incorrectly transcribe audio with speakers with accents or in the presence of background noise, limiting user and context settings. Additionally, ASR makes diarization errors when multiple speakers are present and interrupt one another, such as in a panel discussion where participants argue or get excited. These diarization errors in turn limit the types of conversations ASR works for. 

Currently, the system is designed for two-person audio dialogues and performs particularly well for interviews where one person is creating a majority of the information content and narrative. There are many other types of speech and audio that future work could extend to: political debates, city council meetings, project discussions, doctors appointments, quarterly earnings calls, sporting events, documentaries, and instructions as they all typically share an information dense characteristic. Extending this approach to these areas has many technical challenges including connecting audio to video, increasing ASR accuracy for multi-person audio, and tracing multiple threads of conversation across people.
Lastly, there are also design and privacy considerations around summarizing  personal audio. 
Future work should explore the trade-offs between the benefits of reviewing personal audio and the privacy risks.

\subsection{Summarization Language Model Limitations}
\subsubsection{User Information Retention}
The goal of this system is to help users navigate longform audio and find interesting information quickly which creates user comprehension trade-offs.
A more difficult challenge would be to make a system that helps users find \textit{all} interesting information. We did not measure the precision of the system, but acknowledge that users missed interesting information. In particular, we observe the summarization language model occasionally omits named entities that can signal interest to readers. 
For example, in the interview with hedge fund manager Bill Ackman, he mentions he plays tennis. The system correctly summarizes this fact and several study participants found it interesting. 
The system then produced the \textit{Short Summary} saying "Ackman: He spent the rest of the match trying to hit me back after I hit him with the overhead." 
Many participants overlooked this. 
However, the person Bill Ackman mentions hitting is tennis legend John McEnroe. 
It is likely users  missed this potentially interesting fact due to summarization omission. Future work could experiment with replacing references with corresponding named entities in \textit{Short Summaries} to provide better information scent for users. A first step would be to quantify the interesting information that users missed and then identifying the source: ASR system level errors (transcription, diarization), segmentation errors, or language model summarization errors. Understanding where errors stem from can inform subsequent research where the system should be improved upon to provide more complete information.

\subsubsection{Subtle System Errors May Create User Misunderstanding}
In addition to overlooking information, users may be misled by the system if output summaries contain errors that users do not detect. 
Typically errors are often obvious due to the surrounding context or the reader's prior knowledge of the topic. However, subtle non-obvious errors may go unnoticed, obscuring system inaccuarcies to users. This can lead to misunderstandings as users may not realize a correction is needed to properly understand the underlying information. Our evaluation did not address this limitation. An important next step is to study whether or not the summarization system contains these subtle errors, and if so, how frequent they are. Depending on the nature and severity of such errors, systems could contextually reason entire dialog transcripts (as opposed to individual local semantic segment) or query outside information to check summary adequacy.

\subsubsection{External Knowledge (Co-)Referencing}
Speech summarization is particularly challenging when speakers reference ambiguous objects without proper introduction. Such instances include referencing world knowledge (i.e. current events and facts), local knowledge (speaker dependent context), and deictic references (entity a speaker is directly pointing at). 
In our study recordings, speakers often referenced world knowledge. For example, in the Diversity and Inclusion audio, our system produced the \textit{Short Summary}: "The two children I have look like dante[sic] and rashad." Some users picked up that this referred to Daunte Wright and Rashad Turner, recent tragedies that were foundational to the Black Lives Matters movement, while other users missed this fact. In future work, it may benefit readers to link indirect references in the summaries to outside knowledge.

\section{Conclusion}
Longform spoken dialog is a rich source of information that people encounter every day. However, for individuals who lack the time or inclination to listen to complete audio, the information is not easily accessible.
Given the marked practical performance improvement in ASR and large scale summarization language models, new opportunities and affordances exist that were previously unexplored. These possibilities give hope that content rich longform spoken dialog could be easier to access for individuals who lack the time, inclination, or ability to listen to complete audios. 
However, current state of the art ASR and summarization models still present several sources of error, ranging from word recognition failures during speech recognition to the introduction of hallucinations, grammatical errors, and misrepresented contexts during summarization. Such errors ultimately suggest that immediate fully automated approaches are still out of reach and careful consideration must be given towards creating systems and algorithms to compensate for such shortcomings.
This paper presents an end-to-end dialog summarization system that efficiently repurposes existing ASR and language models while mitigating their innate flaws and present an accompanying user interface that allows for user controllable consumption of hierarchical information.

While prior work has established the benefits of hierarchical browsing of information \cite{zhang2017wikum} \citep{truong2021automatic}, we demonstrate that automatic summarization is now feasible through large scale language models.
Moreover, hierarchical browsing can help people skim information and recover from errors made by automatic tools. Interestingly, we also observe that users can apply \textit{Short Summaries} as a navigational tool to identify interesting parts of audio recordings and drill deeper. From a technical perspective, our system achieves practical summarization accuracy of spoken dialog without custom data sets \cite{karlbomabstractive} or subsequent retraining of large language models. Rather, by using an improved segmentation approach we were able to employ out of the box industry standard models to produce readable and meaningful summaries in a novel traversable interface.

\bibliographystyle{ACM-Reference-Format}
\bibliography{main}

\newpage

\appendix

\section{Algorithm Pseudocode}
\label{app:algos}

\subsection{Coreferenced Semantic Segmentation Pseudocode}
\begin{algorithm}
\caption{Coreferenced Semantic Segmentation}
\label{sec:alg1}
\KwIn{List $\mathbf{T}_{in}$ of $n$ speaker turn segments $\{S\}_{i=0}^{n}$ where $\text{word } w \in \text{sentence } s \in S$, hyperparameters $m=100$ for maximum coreference word span and $p=3$ for minimum number of coreference mentions.} 
$\mathbf{T}_{out} \gets $ list() \textit{\# all semantically segmented speaker turns}\; 
Initialize $e_0, e_f$  \textit{\# coreference span start and end pointers}\;
Initialize $\mathbf{B} \gets $ list(\textit{"I", "me"}) \textit{\# stop tokens}\; 
\For{$i=0,1,...,n$}{
$\mathbf{C} \gets \text{Coreference}(S_i)$ \textit{\# Allen NLP API}\; 
    \For{coreference entity $c \in \mathbf{C}$}{
        \If{c.span $> m$ and $|c| < p$}{ 
            delete $c$ from $\mathbf{C}$\;
        }
        \If{c $\subseteq$ $\mathbf{B}$}{
            delete $c$ from $\mathbf{C}$\;
        }
    }
    $\mathbf{S}_{new} \gets $ list() \textit{\# instantiate new speaker turn block}\;
    $\mathbf{P} \gets $ list() \textit{\# semantic topic cluster within speaker turn} $\mathbf{S}_{new}$\;
    $e_0, e_f \gets c_{min}$ with $min(w_0 \in c \in \mathbf{C})$ \textit{\# entity with earliest word index}\;
    $\mathbf{C}_{used} \gets $ list($c_{min}$)\;
    \For{$s \in S_i$}{
        $s_0, s_f \gets w_0, w_f \in s$ \textit{\# start and end word indices of} $s$\; 
        \For{coreference entity $c \in \mathbf{C}$ and $c \notin \mathbf{C}_{used}$}{
            $c_0, c_f \gets  w_0, w_f \in c$ \textit{\# start and end word indices entity} $c$\; 
            \If{$s_0 \leq e_0$ and $e_f$ > $s_f$}{
                $\mathbf{P}$.append(s) \textit{\# s within span, add to semantic block}\;
                \If{$c_f$ > $e_f$ and $s_0 \leq c_0 \leq s_f$}{
                    $e_0, e_f \gets c_0, c_f$ \textit{\# update maximal entity span}\;
                    $\mathbf{C}_{used}$.append($c$)\;
                }
                break\;
            }
            \Else{
                $\mathbf{S}_{new}$.append($\mathbf{P}$) \textit{\# add topic cluster to speaker block}\;
                $\mathbf{P} \gets $ list(s) \textit{\# begin new topic cluster}\;
                $e_0, e_f \gets c$ with $min(w_0 \in c \in \mathbf{C})$ and $c \notin \mathbf{C}_{used}$\;
                $\mathbf{C}_{used}$.append($c$)\;
                break \;
            }
        }
    }
    $\mathbf{T}_{out}$.append($\mathbf{S}_{new}$) \textit{\# add semantically segmented speaker}\;
}
\Return $\mathbf{T}_{out}$ \textit{\# all semantically segmented speaker turns}
\end{algorithm}

\subsection{Hierarchical Concept Clustering and Merging Pseudocode}

\begin{algorithm}
\caption{Hierarchical Concept Clustering and Merging}
\label{sec:alg2}
\KwIn{List of summaries (sentences) $s \in S_i \in \mathbf{S}$ in speaker turn $S_i$ for all speaker turns $\mathbf{S}$, Embedding \texttt{Sentence Transformer} Model $M$, $cut\_off=5$ determining the smallest concatenated summary allowable}
$L_{out} \gets $ list()\; 
\For{$S_i \in \mathbf{S}$}{
    $\mathbf{E} \gets M{.embed}(s \in S_i)$ \textit{\# embed all summaries within current speaker turn}\;
    $\mathbf{D} \gets \text{create\_pairwise\_cosine\_distance}(\mathbf{E}, \mathbf{E})$\;
    $labels \gets \text{agglomerative\_clustering}(\mathbf{D})$\;
    $cluster \gets [s \in S_i].group\_concatenate(labels)$\;
    $cluster_{filtered} \gets [s > \text{cut\_off} \in S_i]$\;
    $L_{out}.append([cluster_{filtered}])$\;
    }
\Return $L_{out}$
\end{algorithm}
\vspace{0.5cm}

The final level of \textit{Medium} to \textit{Short} summaries contain far fewer summaries then \textit{ASR Transcript} to \textit{Long Summary} and \textit{Long Summary} to \textit{Medium Summary} due to previous merges. We remove the proximity constraint from \textit{Medium} to \textit{Short} and allow agglomerative clustering is across all summaries instead of within speaker turns.

\newpage

\section{Heuristic Score Discussion}
\label{sec:heuristic_score}
\subsection{Heuristic Score Motivation}
In evaluation, a longer reference text (an ASR segment) and a model's generated summary are passed in as inputs to an evaluation model that computes a numerical score describing the summary's accuracy. Usually a text generation's quality estimation focuses on adequacy (content faithfulness) and fluency (coherence). Unlike typical evaluation, our dialog's evaluation setting is unsupervised which poses an extra set of challenges. Our heuristic adopts two individually unsupervised components, \texttt{BERTScore} and \texttt{Sentence Transformer} to measure adequacy (information retention, \textit{or semantic content}) and to a lesser degree fluency (coherence \textit{or grammaticality}). Evaluating a generated summary's fluency without a reference is notably challenging (discussed below) but is, to some extent captured, within \texttt{BERTScore}'s token embedding matching.

\subsection{Heuristic Score Trade-offs}
Unsupervised systematic and automatic evaluation of a summarization model's text generation is particularly difficult as it requires comparing generated sentences to non-existent annotated references. The benefit of the unsupervised reference-free evaluation setting of the heuristic score fundamentally incurs similar trade-offs to that of the underlying automatic evaluation methods \texttt{BERTScore, Sentence Transformer}.

Because of the lack of human authored summaries in both training and evaluation data, we are unable to use standard summarization evaluation standards such as ROUGE (Recall-Oriented Understudy for Gisting Evaluation) \cite{lin2004rouge}. Moreover, existing $n$-gram evaluation metrics cannot be effectively used on summaries as they tend to fail at robustly capturing performance; shortened paraphrases containing semantically critical changes in word positions may be unfairly penalized due to a mismatch with the reference text. 

One subsequent trade-off is the possibility of achieving adequacy at the cost of low fluency. Recall that semantic content is compared to an entire (unsummarized) dialog segment as a reference; thus a model can theoretically fool the metric by outputting many \textit{shorter} and \textit{less relevant} segments in an incoherent manner instead of a single relevant segment to achieve a better score, as is the case with \textbf{\texttt{M2}}'s marginally better score over \textbf{\texttt{M3}}'s. To ensure that this is not the case, we also qualitatively inspect our formative study's findings in Section 3.4.

\end{document}